# Loop Improvement: An Efficient Approach for Extracting Shared Features from Heterogeneous Data without Central Server

Fei Li, **Chu Kiong Loo,** *Senior Member, IEEE,* Wei Shiung Liew, Xiaofeng Liu

**Abstract**—In federated learning, data heterogeneity significantly impacts performance. A typical solution involves segregating these parameters into shared and personalized components, a concept also relevant in multi-task learning. Addressing this, we propose "Loop Improvement" (LI), a novel method enhancing this separation and feature extraction without necessitating a central server or data interchange among participants. Our experiments reveal LI's superiority in several aspects: In personalized federated learning environments, LI consistently outperforms the advanced FedALA algorithm in accuracy across diverse scenarios. Additionally, LI's feature extractor closely matches the performance achieved when aggregating data from all clients. In global model contexts, employing LI with stacked personalized layers and an additional network also yields comparable results to combined client data scenarios. Furthermore, LI's adaptability extends to multi-task learning, streamlining the extraction of common features across tasks and obviating the need for simultaneous training. This approach not only enhances individual task performance but also achieves accuracy levels on par with classic multi-task learning methods where all tasks are trained simultaneously. LI integrates a loop topology with layer-wise and end-to-end training, compatible with various neural network models. This paper also delves into the theoretical underpinnings of LI's effectiveness, offering insights into its potential applications. The code is on https://github.com/axedge1983/LI

**Index Terms**—Federated Learning, Multi-Task Learning, Shared feature extraction, Layer-wise training

—————————— ◆ ——————————

## 1 INTRODUCTION

FEDRATEDS Learning (FL)[1] is a distributed machine learning approach that enables model training on decentralized devices or data sources. It prioritizes data privacy while facilitating knowledge sharing. However, when FL is confronted with non-IID (non-Independently Identically Distributed, a form of data heterogeneity that we will not differentiate in this paper) data, significant challenges arise. In such scenarios, local optimization goals may not align with global ones, potentially leading to degraded performance [2].

To address this issue, researchers have introduced Personalized Federated Learning (PFL). PFL aims to provide personalized models for each client and improve convergence on highly heterogeneous data [2]. The strategies employed in PFL involve initial training of a global model, followed by fine-tuning it on each client or directly decoupling local private model parameters from global FL model parameters, bearing similarities to Multi-Task Learning (MTL) [3]. PFL can enhance the performance of models owned by each company without disclosing the company's own data, thereby motivating companies to share data.

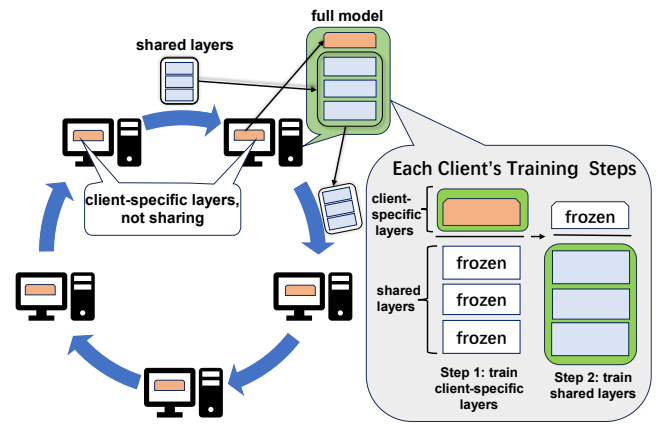

Fig. 1. Left: LI method in FL, with shared layers and client-specific layers. Right: An execution instance for a client.

In MTL, the goal is to train a model capable of performing several related tasks simultaneously, enhancing task generalization and reducing overfitting risks [4]. "Even when dealing with the same dataset, the data can be considered heterogeneous, as the labels for each task vary. In multi-task learning, even with the same dataset, different tasks may focus on different features and patterns. This variation in task objectives, even with identical data, represents a form of data heterogeneity. For example, in autonomous driving tasks, different sub-tasks (such as detecting traffic lights, road markings, and other vehicles) require the model to focus on different features within the same image data. MTL

————————————————

- *Fei Li is with the Faculty of Computer Science and Information Technology, Universiti Malaya, Kuala Lumpur, Malaysia, CO 50603 & Faculty of Artificial Intelligence, Neijiang Normal University, Neijiang, China, CO 641100. E-mail: S2021084@siswa.um.edu.my & 21021312@zju.edu.cn.*
- *Chu Kiong Loo is the corresponding author, he is with the Faculty of Computer Science and Information Technology, Universiti Malaya, Kuala Lumpur, Malaysia, CO 50603. E-mail: ckloo.um@um.edu.my.*
- *Wei Shiung Liew is with the Faculty of Computer Science and Information Technology, Universiti Malaya, Kuala Lumpur, Malaysia, CO 50603. E-mail: liew.wei.shiung@um.edu.my.*
- *Xiaofeng Liu is with the Department of Telecommunication, HoHai University, Nanjing, China, CO 213200. E-mail: xfliu@hhu.edu.cn.*

typically shares parts of the underlying layers across all tasks while preserving several task-specific output layers. When dealing with scenarios where different data distributions reside on different clients and these data are usually heterogeneous, this scenario is referred to as Federated Multi-Task Learning (FMTL) [7]. In the context of FMTL, each FL client can be considered a distinct task in MTL, allowing the extraction of shared features despite the heterogeneity of clients' local data.

Both MTL and FL typically avoid sequentially training tasks/clients. This is particularly critical when dealing with non-IID data. Due to its inherent dissimilarity and irregularity, non-IID data can exacerbate the problem known as 'catastrophic forgetting' [21]. This common phenomenon in artificial neural networks refers to the abrupt loss of knowledge from previously learned task(s) (e.g., task A) as the model incorporates information relevant to the current task (e.g., task B). Notably, studies from Ramasesh [17] and Luo, M. [18] indicate that the 'model head' - the final layers of a neural network where task-specific learning occurs - plays a crucial role in the occurrence of catastrophic forgetting. Their findings served as a key inspiration for our research.

Inspired by recent studies, we propose a simple method called Loop Improvement (LI) under the domain of FL and MTL. This method introduces a loop topology where each node (a client or a task) possesses its own unique personalized layers (the 'head'). Additionally, all nodes share a common part consisting of some layers (the 'backbone'). LI is a method that combines layer-wise training from higher to lower layers with end-to-end training, and it can be used in combination with any neural network model. The training protocol for each client (task) is divided into two distinct steps: the initial training of personalized layers, followed by the training of shared layers. The parameters of the shared layers (excluding the personalized layers) then transferred to the next node. This process loops until it reaches a predetermined round limit. This approach effectively decouples shared and personalized parameters. In the initial step, the 'head' is trained while the 'backbone' is kept frozen. This allows for an initial understanding of the unique characteristics of each node, and importantly, mitigates the issue of catastrophic forgetting by reducing the discrepancy caused by varied supervisory signals from the top layers towards the shared layers. In the second step, the 'backbone' learns universal attributes across tasks (clients), identifying features that can enhance performance in the upper layers, leading to efficient shared base layers. After mitigating the effects of data heterogeneity, the data for each node within this loop can be viewed as forming a larger batch. The training process on a particular node can be regarded as updating the model parameters for one batch, and one round of learning in the loop is equivalent to one epoch, where the labels are supervisory signals backpropagated from the upper layers. In this view, the training of model parameters includes only the shared layers and excludes the personalized parts. Fig. 1 shows the principle of LI method. A more detailed explanation of why the LI method is effective will be expounded in the methods section. For scenarios requiring a global model, the training steps at each node can further include an optional step, as depicted in Fig. 2.

The mainstream architecture of FL relies on a central server to aggregate parameters from each client. However, there are also decentralized Federated Learning methods that enable direct communication between clients. The LI method falls under the category of decentralized Federated Learning methods, operating in a peer-to-peer manner, thus eliminating the need for a central server. This algorithm shares model parameters instead of raw data. As such, it meets the basic privacy protection requirements in FL scenarios. In scenarios where a global model is necessary, we have proposed several solutions in this paper. Among them, stacking outputs from each personalized layers and training an auxiliary network has proven effective for achieving a global model.

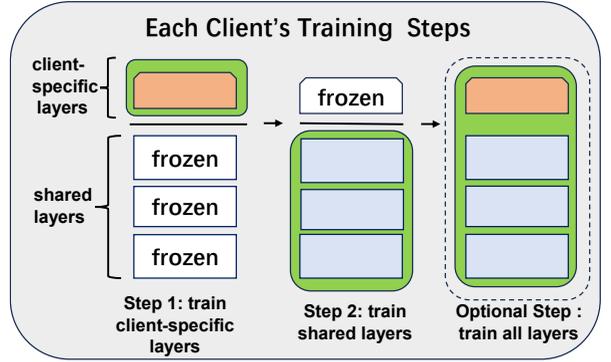

Fig. 2. Optional Step Added to Each Node for Scenarios Requiring a Global Model

The LI method, adopting a loop structure, is better suited for cross-silo federated learning situations [5] (clients are data centers distributed across different organizations or geographical locations) demanding stable networks and clients across silos, rather than a more general cross-device federated learning scenario. For unstable network scenarios, we also proposed a solution using a dual-loop redundant structure, like FDDI [25]. Fig. 3 shows its principle.

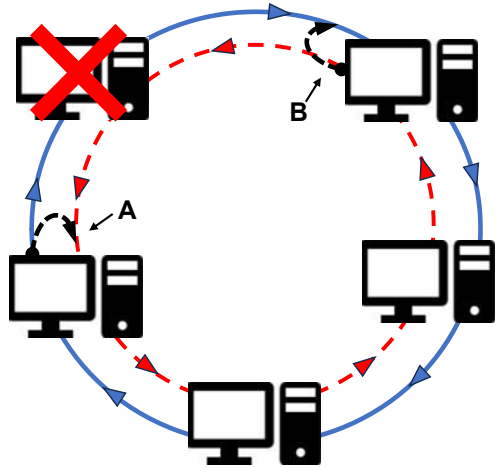

Fig. 3. Dual-Loop Redundancy: The outer loop operates normally. In case of a disruption, the inner loop activates, with Points A and B re-establishing the loop.

At first glance, the training process for each node may seem sequential. However, after the initial node completes training for one or a few epochs, it can send the shared layers to the next node, after which both the initial and subsequent nodes can immediately start training. Thus, the LI method facilitates a loop pipeline structure amongst nodes, enabling parallel training. This approach unifies MTL and FL. From the perspective of LI, there is no fundamental distinction between FL and MTL, as both involve loop training and include several personalized components along with a shared part, as depicted in Fig. 4. While FL and MTL typically steer clear of sequential training for each task due to the risk of catastrophic forgetting, LI method demonstrates the effectiveness of a loop pipeline approach when combined with a two-step training process.

The LI method is actually a combination of layer-wise training [37] from top to bottom and end-to-end training. In the early stages of deep learning, layer-wise training from the bottom to the top was very popular. With the development of GPU computing technology, end-to-end training of neural networks has become more common. We have divided the neural network into two parts, and conduct training from the top to the bottom, achieving great results in federated learning. To our knowledge, this is the first time this idea has been applied to federated learning and multi-task learning. Our contributions can be summarized as follows:

1. We introduce the 'Loop Improvement' (LI) method, a novel approach in Federated Learning (FL) and Multi-Task Learning (MTL) that negates the need for data sharing or a centralized server. This method excels in extracting shared features across diverse contexts. While simple in its implementation, LI demonstrates remarkable effectiveness, outperforming existing methods as evidenced by our comprehensive experiments. These experiments also highlight the LI method's ability to mitigate the effects of data heterogeneity through initial head training. Particularly in Personalized Federated Learning, LI significantly boosts performance across all clients, encouraging data sharing. The methodology section provides a detailed rationale for the efficacy of this approach.

2. We present a straightforward yet efficient framework based on the LI method for developing a global model in general FL scenarios. Our experiments, conducted across multiple datasets, establish that the LI method achieves accuracy levels comparable to those obtained by training with combined data from all clients. Additionally, this paper outlines tailored solutions to address client and network failures within the LI method's loop structure, enhancing its robustness and applicability in real-world scenarios.

3. We explore the potential future applications of the LI method, envisioning its adaptability across various datasets, modalities, and task types. This forward-looking perspective underscores LI's versatility and its promising role in advancing the fields of Federated Learning and Multi-Task Learning.

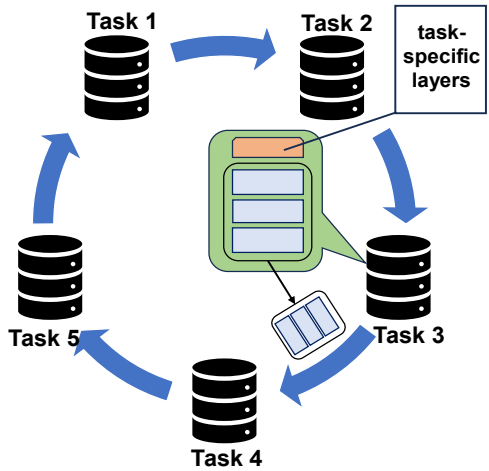

Fig. 4. LI in MTL: Demonstrating a similarity with its use in FL. Data for multiple tasks can be hosted on the same or different machines, and each task has its own task-specific layers.

## 2 RELATED WORKS

Federated Learning (FL) was introduced by Mcmahan et al. in 2017 [1]. The initial FedAvg method aggregates a global model from individual client models and is highly reliant on IID data distribution [1]. Nevertheless, many real-world applications encounter non-IID data distributions. To address this, Personalized Federated Learning (PFL) emerged, aimed at developing personalized models for individual clients, considering their unique data distributions. One effective strategy adopted in PFL is parameter decoupling, which separates local private model parameters from the global FL model parameters, allowing for the learning of specific task representations and enhancing personalization [2]. The current newer and more effective method in PFL, however, is the FedALA method [20]. In their experiments, FedALA surpassed more than a dozen of the current mainstream methods. Several FL strategies have been proposed to manage the distinction between shared and personalized layers. For instance, the FedPer algorithm allows for central training of the base layer through FedAvg while local training is applied to the personalized layers [3]. Meanwhile, FedWeIT introduces a federated continual learning method that decomposes network weights into global federated parameters and sparse task-specific parameters, thus facilitating knowledge transfer between clients during the learning process [6].

The FedBABU algorithm aggregates only the 'body' part of the model, which is associated with universality, during federated training. Subsequently, the 'head' part of the model, which is associated with personalization, is fine-tuned. This distinction reveals that improvements in the global model performance do not necessarily translate into enhanced personalization performance [8].

The LG-FEDAVG model decomposes model parameters into global feature vectors and local features, with each local device learning how to extract higher-level representations from raw data, followed by global model learning on these representations [9]. The FedRep algorithm learns a global low-dimensional representation using data from all clients, and each client computes a personalized, low-dimensional classifier, referred to as the client's "head" [10].

Alternating training in MTL can be seen as a loop structure. To maintain model information during such alternating training, Kim et al. utilized a method known as Learning without Forgetting (LWF), based on knowledge distillation [11]. Similarly, the MultiCoFusion framework adopts a multi-task alternating learning strategy, initially using a pre-trained network to separately learn feature representations of histopathological images and mRNA expression data. These representations are then fused together using a fully connected neural network, forming a shared multi-task network [12].

AdaMTL provides a complex adaptive framework consisting of an encoder, multiple task-specific decoders, and a lightweight policy network. Initially, the encoder and decoders are trained. They are then frozen to fine-tune the policy network. Ultimately, the policy network and decoders undergo alternating training [13]. The policy network actually adopts the MoE (Mixture of Expert) idea [14].

As for the application of the loop structure in FL, the CWT method applies this structure to FL with the goal of enhancing the performance of deep learning algorithms in medical image diagnosis. It enables the periodic transfer of the model's weights among different institutions, facilitating the process of knowledge sharing and transfer [15]. According to a study conducted by L. Qu et al., the Vision Transformer (ViT) [23] structure is found to be particularly compatible with heterogeneous data. By merely replacing CNNs with ViTs, it was demonstrated that both CWT and FedAvg could maintain model accuracy even in highly heterogeneous non-IID settings [16].

In the realm of head-first training, studies conducted by Ramasesh et al. showed that tasks of medium similarity suffer the most from forgetting, a phenomenon primarily driven by higher layers in the model. Interestingly, they discovered that pre-training the new task's head for a few epochs before training the whole network can alleviate the performance drop on the original task [17]. Echoing this, Luo et al. observed a larger bias in the classifier compared to other layers. Consequently, they proposed the CCVR algorithm, which mitigates this bias by sampling virtual representations from an approximate Gaussian mixture model [18]. It is observed by Nguyen et al. that using a pre-trained model initialization can reduce the effect of system heterogeneity [22]. In their work, Legate, G. et al. a two-phase approach of first obtaining the classifier and then fine-tuning the model can yield rapid convergence and improved generalization in the federated setting. The first step involves FL to obtain a classification head (Head-Tuning stage), followed by an extensive fine-tuning process (Fine-Tune stage) to generate the global model, they show that fitting a classifier using the Nearest Class Means (NCM) can be done exactly and orders of magnitude more efficiently than existing proposals, while obtaining strong performance [19].

## 3 METHOD

### 3.1 Steps and Details of the LI Algorithm

As illustrated in Fig. 1, we initially segment the network into the personalized layers and the shared layers. In our experiments, using the Cifar10/Cifar100 dataset [26] and Tiny-ImageNet [34] as examples, a fully connected layer is adequate to serve as the personalized layers. For more complex datasets and models, a more refined separation of shared and personalized layers may be necessary, possibly even dividing them into three or more parts. Each node (in the context of FL, it is a client, and in MTL, it represents a task) possesses its personalized layers, with the personalized layers of MTL corresponding to their respective tasks. The entire loop structure has only one shared layer, initially located at the initial node.

Subsequently, starting from the initial node, we train each node through the following two steps:

1. **Step 1 (Personalized Layers Training):** At this stage, we freeze the parameters of the shared layers and train only the parameters of the node's personalized layers.
2. **Step 2 (Shared Layers Training):** Then, we freeze the parameters of the personalized layers and train only the parameters of the shared layers.
3. **Optional Step (All Layers Training):** Specifically applicable in scenarios where acquiring a global model is crucial, incorporating this step can slightly enhance the accuracy of the global model. Here, we release all parameters from their frozen state, enabling the training of both shared and personalized layers in unison, as shown in Fig. 2.

Upon completion of the training process for the current node, we then transfer the parameters of the shared layers (excluding the personalized layers) to the next node. Following this, we move to the next nodes in the sequence. This iterative process continues until all nodes have undergone the training sequence, marking the end of a round. This loop is repeated until the maximum preset number of rounds is reached. The procedure of LI is outlined in Algorithm 1.

The description of the parameter update process is as follows:

Each client possesses an independent head $\theta_{\text{head}_c}$, which enables the model to adapt locally to the specific data distribution of each client. The update process for the head can be represented as:

$$\theta_{\text{head}_c}^{new} = \theta_{\text{head}_c} - \quad (1)$$

$$\eta \nabla_{\theta_{\text{head}_c}} L_c \big( g\big( f(x_c; \theta_{\text{backbone}}^{fixed}), \theta_{\text{head}_c} \big), y_c \big)$$

Where $x_c$, and $y_c$ are the data and labels of client $c$, respectively, $\eta$ is the learning rate, and $L_c$ is the loss function of client $c$. Here, $g$ represents the head network, and $\theta_{\text{backbone}}^{\text{fixed}}$ indicates that the backbone parameters remain fixed during the head update. This arrangement allows each client's head to be specifically adjusted to its own dataset, thus better addressing the non-IID characteristics of the data. The backbone $f(x; \theta_{\text{backbone}})$ is responsible for feature extraction. In a loop structure, each client $c$ updates the head $\theta_{\text{head}_c}$ and then updates the backbone $\theta_{\text{backbone}}$, which is then passed on to the next client. This update process can be represented by the following formula:

$$\theta_{backbone}^{new} = \theta_{backbone} - \eta \nabla_{\theta_{backbone}} L_c(f(x_c; \theta_{backbone}), \theta_{head_c}, y_c) \quad (2)$$

In this way, the backbone learns and extracts features, and undergoes adaptive adjustments before being passed on to the next client.

---

**Algorithm 1:** Loop Improvement (LI)

**Data:** Total number of clients $C$, data loaders $\{D_c\}$ for each client $c$

**Result:** Updated backbone parameters $\theta_{\text{backbone}}$, head parameters $\{\theta_{\text{head}_c}\}$ for each client $c$

1 Initialize shared backbone parameters $\theta_{\text{backbone}}$;

2 **for** *each client $c$ in 1 to $C$* **do**
3     Initialize head parameters $\theta_{\text{head}_c}$;
4 **end**

5 **Function** *Train_client ($c$, $\theta_{backbone}$, $\theta_{head_c}$, $D_c$, $E_{head}$, $E_{backbone}$, $E_{full}$)*:
    // Train head while backbone is frozen
6     Freeze parameters in $\theta_{\text{backbone}}$;
7     Unfreeze parameters in $\theta_{\text{head}_c}$;
8     Train $\theta_{\text{head}_c}$ for $E_{\text{head}}$ epochs using $D_c$;
    // Train backbone while head is frozen
9     Unfreeze parameters in $\theta_{\text{backbone}}$;
10     Freeze parameters in $\theta_{\text{head}_c}$;
11     Train $\theta_{\text{backbone}}$ for $E_{\text{backbone}}$ epochs using $D_c$;
    // Optional: Train full model
12     Unfreeze all parameters;
13     Train $\theta_{\text{backbone}}$ and $\theta_{\text{head}_c}$ for $E_{\text{full}}$ epochs using $D_c$;
14 **return**

15 **for** *round in 1 to $R$* **do**
16     **for** *each client $c$ in 1 to $C$* **do**
17        $\theta_{\text{backbone}} \leftarrow$ Receive backbone parameters from previous client;
18        Train_client($c$, $\theta_{\text{backbone}}$, $\theta_{\text{head}_c}$, $D_c$, $E_{\text{head}}$, $E_{\text{backbone}}$, $E_{\text{full}}$);
19        Send $\theta_{\text{backbone}}$ to the next client;
20     **end**
21 **end**

### 3.2 Why is the LI Method Effective?

Let's now discuss why the LI method is effective. We first look at the process of training the shared layers under the LI method from a more macro perspective:

All the data from the nodes are viewed as a large-scale dataset, with each node's data being considered one batch of this larger dataset (with the batch size being the size of the data each client possesses). Each training process on a node can be thought of as one batch update of the model parameters. A single loop, where the shared layers pass through each node once, is considered as one large epoch. We view this entire loop as a single epoch in the context of the LI method. In this context, the supervision signal for the model (shared layer) no longer comes directly from the original labels, but from the gradient information back-propagated from the personalized layers. Thus, the target model for training is no longer the entire model, but only includes the shared layers that every node will update.

Taking the situation of non-IID labels as an example: we first train the personalized layers, which not only adapt to the mapping between labels and the output from the shared layers, but also help to mitigate the impact of catastrophic forgetting on the shared layers as the personalized layers have already handled the parts with larger differences. In the second step of training the shared layers, the input comes from the original data, and the supervision signal comes from the backpropagation of the personalized layers. The shared layers need to adapt to the input data from all nodes and the supervision signals from all nodes, thus being forced to learn the common features of all data.

In the context of the LI method, a training round is seen as an epoch, and the number of rounds is regarded as the number of epochs. A single training iteration may not significantly influence the shared layers, but through multiple rounds of training across multiple clients, the shared layers can ultimately learn the common characteristics of all data.

Next, we examine the process of training the personalized layers under the LI method from a more macro perspective:

For a certain node n, the input data is no longer the original data, but the features output by the current shared layers; the labels remain the original labels. In this scenario, the target of training is no longer the entire model but just the personalized layers of node n. The personalized layers need to adapt to the supervision signal from the labels of node n and the "data" from the current shared layers. The shared layers extract a powerful shared feature representation from all the data, and the personalized layers adapt to this feature and the unique labels of node n based on this. After several rounds of training, the shared layers become increasingly robust, correspondingly, node n can also obtain a better personalized layers model.

The training of the personalized layers and the shared layers is a step-by-step process. They complement each other and progress together, ultimately leading to shared layers that can extract the feature representation of all data and the personalized layers that is compatible with both the node labels and the shared layer features.

In contexts where a global model is desired, we have observed that including the additional optional step can enhance the global model's accuracy. This improvement may be attributed to the optional step's ability to facilitate more effective collaboration between the shared and personalized layers.

## 3.3 Flexible Applicability of the LI method

The LI method possesses significant flexibility, capable of adapting to various learning scenarios. Thus far, we mainly discussed scenarios involving label distribution differences. Regardless of differences in the underlying data distribution or more complex learning environments, we can train the shared layers and personalized layers independently, following the previously defined steps 1 and 2. This is because, in any case, the shared layers and personalized layers need to adapt to all signals from the upper and lower layers.

The LI method demonstrates remarkable adaptability, allowing for easy diverse configurations. For example, we can construct more shared layers and fewer personalized layers as needed, or vice versa. We could even place the personalized layers at both ends, with the shared layers in the middle, or vice versa. If we hope for larger changes in certain personalized layers in a specific task, yet they exhibit small changes in reality, we can increase the number of personalized layers training times during each node training.

For tasks that require more precise personalized models, after obtaining the shared layers, we can perform several rounds of 'head' fine-tuning to optimize the performance of these models.

It is worth noting that the LI method does not replace other methods but serves as a new training strategy that can be combined with other methods, providing more training choices and possibilities. LI can also be paired with any neural network model.

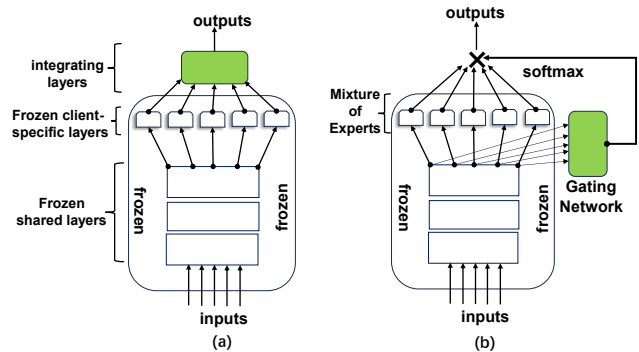

Fig. 5. (a) Process of obtaining a global model by stacking all personalized layers, where training includes directly obtaining prediction results from the integrating layers. (b) Process of obtaining a global model using the MoE method, where training involves the gating network to determine the weight of each expert.

## 3.4 Global Model Generation

In the context of FL, if there is a need to generate a global model, we offer three possible solutions:

1. Solution 1: Choose a very small batch for training. After updating the model using a small batch each time, immediately transmit the model to the next client. This method, like small batch training on the entire dataset, can achieve effects close to training on all datasets, and may even bypass the two steps. However, this requires transmitting many model parameters, as each batch needs a model transmission.

2. Solution 2: Stack the trained shared layer and each node's personalized layers. The data first pass through the shared layer and all the personalized layers, and the resulting prediction results serve as the input for a new network (integrating layers), with the output being the real labels. Under non-independent and identically distributed data, we can let the integrating layers execute loop updates on a small batch to train these integrating layers, transmitting the integrating layers' model. Another way is to directly centralize all the output data samples of each client on a single client through a loop method, followed by shuffled training, transmitting each client's prediction results. Whether transmitting the integrating layers' model or each client's prediction results, the original data will not be leaked. Fig. 5. (a) demonstrates this approach.

3. Solution 3: Leverage the classic "Mixture of Experts"[27] strategy, where each client's personalized layers acts as an "expert". An additional gating network is trained to determine the weight each "expert" should contribute. The final prediction is a weighted sum of all "experts"' predictions, exploiting the distinct expertise of each client's personalized layers to boost model performance. Fig. 5. (b) demonstrates this approach.

## 3.5 Parallel Processing and Data Transmission

Despite the seemingly sequential training process, the actual operation is such that after the initial node completes training for one or a few epochs, it can send the shared layers to the next node. At this point, both the initial and the subsequent nodes can start training simultaneously. This creates a structure resembling a loop pipeline, where multiple nodes are engaged in training concurrently. However, this approach increases the communication overhead and requires a careful balance. The focus of this paper is not on how to balance parallel training with communication costs, or on how to implement parallel training more effectively. These considerations are left for our future research.

Compared to the conventional FedAvg, the LI method only needs to transmit half the data volume in the same round. However, when there is a significant discrepancy in the training speed or data volume across clients, the bottleneck of the entire network may occur at the slowest client or the one with the largest amount of data, as all clients need to await the completion of the slowest client. In such scenarios, coordination is necessary. More frequent training of the faster nodes and less frequent training of the slower nodes should be implemented to synchronize the training speed across the network as much as possible, thus maximizing the throughput of the entire loop.

In situations with unstable networks, we propose a potential solution: adopting a dual-loop structure like the Fiber Distributed Data Interface (FDDI), as depicted in Fig. 3. Under normal circumstances, the outer loop route is responsible for data transmission. When a client or network failure occurs (top left corner of Fig. 3), the inner loop route

TABLE 1
PERFORMANCE COMPARISON: FEDLOOP VS. FEDALA

| Dataset | Model | Clients | Heterogeneity | FedALA (SGD) | FedALA (adamW) | LI(adamW) |
|---|---|---|---|---|---|---|
| MNIST | CNN (4 layers) | 5 | Pathological | 99.87 | 99.93 | **99.94** |
| Cifar10 | CNN (4 layers) | 5 | Pathological | 92.16 | 92.38 | **92.59** |
| Cifar10 | CNN (4 layers) | 20 | dir=0.1 | 90.69 | 91.19 | **92.12** |
| Cifar10 | Resnet18 | 5 | Pathological | 91.19 | 92.73 | **93.32** |
| Cifar10 | Resnet18 | 20 | dir=0.1 | 89.05 | 90.9 | **92.5** |
| Cifar100 | CNN (4 layers) | 10 | Pathological | 67.96 | 72.74 | **72.88** |
| Cifar100 | CNN (4 layers) | 20 | dir=0.1 | 57.01 | 58.43 | **58.87** |
| Cifar100 | Resnet18 | 20 | dir=0.1 | 53.93 | 58.94 | **66.64** |
| Cifar100 | Resnet18 | 10 | Pathological | 65.69 | 73.07 | **76.44** |
| Tiny-Imagenet | Resnet18 | 20 | dir=0.1 | 42.4 | 45.94 | **55.17** |
| Tiny-Imagenet | Resnet18 | 20 | Pathological | 52.07 | 59.88 | **64.8** |
| AG-News | FastText | 20 | dir=0.1 | 96.16 | 97.53 | **98.75** |

*The "clients" column indicates the number of clients. "dir=0.1" represents the $\beta$ value is 0.1 for the Dirichlet distribution. "FedALA (adamW)" and "LI (adamW)" denote the use of the adamW optimization method with a fixed step learning rate decay, decaying by 0.5 every 10 steps. The values in the last three columns all represent the percentage of accuracy.*

comes into action (Places indicated by points A and B in Fig. 3), and together with the outer loop route, forms a new circuit. If the network breaks into two parts, the network will split into two subnetworks. However, once the network restores, loop training can be resumed.

## 4 EXPERIMENTS

### 4.1 Personalized Federated Learning

We utilized the environment provided by FedALA (available at: https://github.com/TsingZ0/PFL-Non-IID) and modified their FedAvg code to implement the LI method. We designated only the last linear layer as the head, while the rest functioned as the backbone. In line with the original authors, we adhered to the parameters set out in their paper, using datasets like MNIST [31], Cifar10 and Cifar100 with a 4-layer CNN. In the domain of natural language processing, we explored text classification tasks on AG News [32] and fastText [33]. Furthermore, we utilized ResNet-18 for Cifar10, Cifar100, and Tiny-ImageNet. The local learning rate was set at 0.005, batch size at 10, and the number of local training epochs at 1. We conducted 150 iterations of FedALA empirically. In the LI method, we use only the final fully connected layer as the personalized layers. The initial learning rate for the head is set to 0.0001, while the learning rate for the backbone part is set to 0.0004. All other parameters are exactly the same as those in FedALA.

To simulate heterogeneous settings, we adopted two scenarios. The first was pathological heterogeneity [35], sampling 2/2/10 classes for MNIST/Cifar10/Cifar100 respectively from a total of 10/10/100 classes per client, ensuring disjoint data samples. The second scenario involved practical heterogeneity [36], controlled by the Dirichlet distribution, where smaller 'dir' values indicate greater heterogeneity. We set 'dir' to 0.1 and configured 20 clients.

In PFL setting, we evaluated the learned model on the client side, using 25% of local data for testing and the remaining 75% for training. We implemented the algorithms using PyTorch 2.0 and conducted experiments on a computer with an Intel 10900 CPU, 64G of memory, and an NVIDIA 3090 GPU, running Ubuntu 20.04.

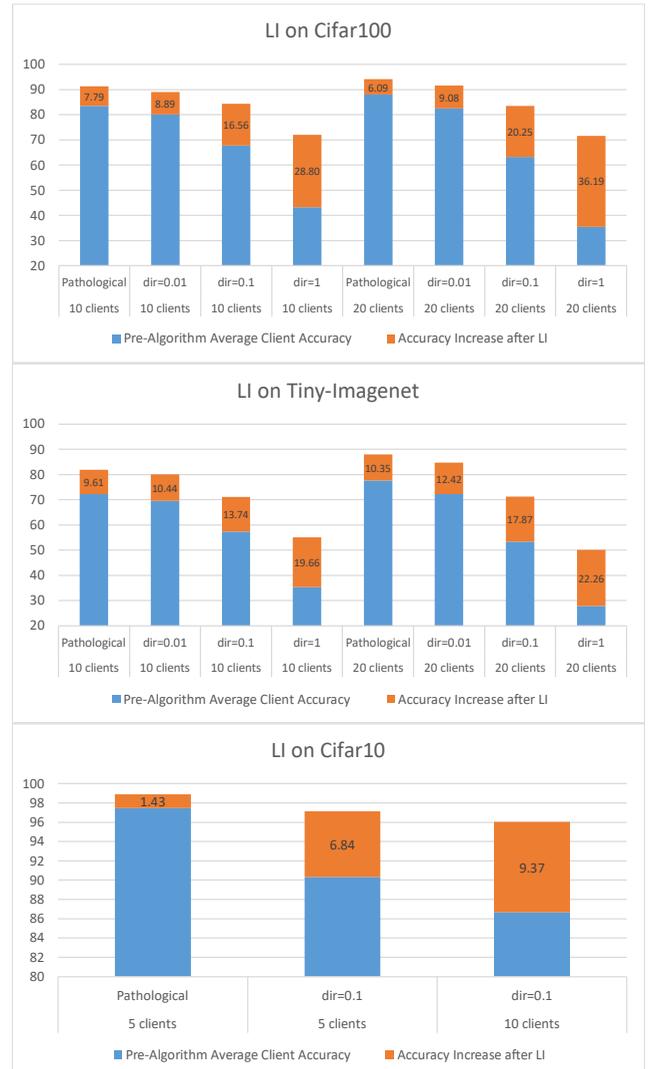

Fig. 6. Accuracy Improvement for Each Client after Application of LI Algorithm. The numerical values represent the percentage of accuracy.

The results of the experiment are shown in Table I, we replicated the authors' accuracy with few differences. We found that by changing SGD to AdamW (with weight decay set to 0.1) and applying lr_step (fixed step learning rate decay) in Pytorch, decaying 0.5 every 10 steps, the capabilities of FedALA were further enhanced. Comparing the LI algorithm with both FedALA+SGD and FedALA+AdamW+lr_step, the LI algorithm outperformed FedALA across all environments and datasets from the authors' paper, regardless of the optimizer (SGD or AdamW) and network architecture (4-layer CNN or ResNet-18).

We then evaluated the improvement in accuracy for each client after applying the LI algorithm. First, we established a baseline by training each client independently (using only their data) to obtain an average accuracy, utilizing the CoAtNet [28]. After running the LI algorithm, we calculated the average accuracy for each client and compared it with the baseline, observing significant improvements, akin to training on a larger dataset for each client. The experimental results, as illustrated in Fig. 6, show a significant improvement in accuracy. It can be observed that the lower the data heterogeneity, the greater the increase in accuracy.

### 4.2 Multi-Task Learning

For multi-task learning tasks, we utilized the CelebA dataset [30]. Initially, we applied a classic multi-task learning approach using a ResNet-18 model with a modified fully connected layer, comprising 40 nodes corresponding to the 40 attributes of the CelebA dataset. We used BCEWithLogitsLoss for the loss function. The accuracy obtained on the test set served as our baseline. The batch size was set to 32.

Subsequently, we divided the dataset evenly into 40 parts, with each part dedicated to a specific task. These tasks were trained separately, and their test accuracies were computed and averaged to yield the Pre-Algorithm Average Accuracy.

Next, we implemented the LI algorithm, where the head of each task was composed solely of a single linear layer. After testing, we averaged the accuracies to obtain the Average Accuracy after LI.

The experimental results, as illustrated in Fig. 7, show that the accuracy of the LI algorithm generally lies between the two metrics. We also tested the LI algorithm without splitting the dataset, which resulted in an accuracy very close to the baseline (91.67% vs. 91.72%).

The experimental results show that the LI algorithm is indeed capable of extracting common features across multiple tasks. In the context of federated multi-task learning, the LI algorithm can improve the accuracy of each client.

### 4.3 Global Model and Shared Feature Extractor

Next, we evaluated the accuracy of the "shared features extracted using the LI algorithm" and the "global model obtained using the LI algorithm." We randomly mixed all client datasets to form a large dataset for training, using the accuracy of this model as our baseline. We then compared the accuracy of the shared features and the global model, both obtained through the LI algorithm, against this baseline.

To assess the effectiveness of the "shared features extracted using the LI algorithm," we froze the shared layers of the model. Training was then conducted on the large dataset created by mixing all client datasets. The head used for training was identical to that of a particular client, with only the head being trained. The performance was compared with the baseline accuracy to verify effective feature extraction.

For acquiring the global model, we followed the method depicted in Fig. 5. (a), integrating all client-specific layers with a simple integrating layer (new network). We fixed all shared and personalized layers, training only this integrating layer. This layer's input size is minimal, making it feasible in real-world scenarios to save and mix feature maps from all datasets after passing through shared and personalized layers, thus training on feature maps from all clients without data leakage. For the Cifar10 dataset, due to the small size of feature maps, the personalization layer included both a linear layer and the last layer of the last transformer module, removing the fully connected layer from the personalization layer to obtain the global model (as the integrating layers have a fully connected layer).

In this part of the experiment, we used the same division for both training and test sets as in FedLab [29]. For the training set, we applied data augmentation techniques in Pytorch: RandomHorizontalFlip(p=0.5), RandomCrop(padding=4), TrivialAugmentWide (BILINEAR),

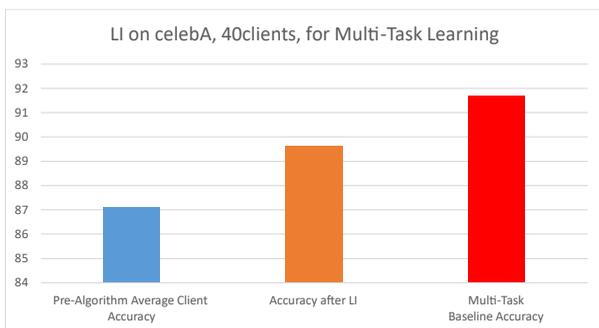

Fig. 7. Application of LI Algorithm in Multi-Task Learning on the CelebA Dataset. The values represent the percentage of accuracy.

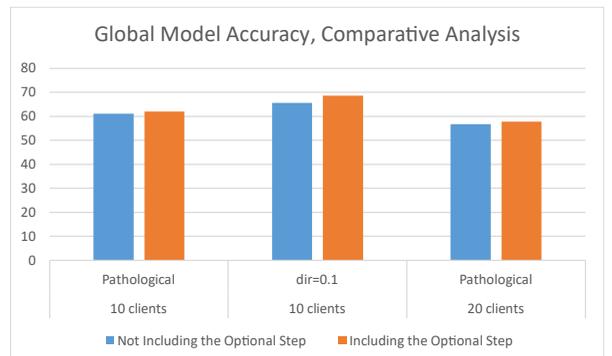

Fig. 8. Comparative Analysis: Strategies with and without the Inclusion of the Optional Step. The approach including the optional step is slightly better than the one without it. The values represent the percentage of accuracy.

RandomErasing(p=0.1). No data augmentation was used for the test set.

We employed the CoAtNet [28] model for these experiments. For the implementation of CoAtNet, we used this source(https://juliusruseckas.github.io/ml/coatnet.html) on Cifar10 and Cifar100, and this source (https://github.com/chinhsuanwu/coatnet-pytorch) on Tiny-ImageNet. The batch size was set to 100 for all. We used AdamW (with weight decay set to 0.1), with an initial learning rate of 0.0005, a head learning rate of 0.0001, and a backbone learning rate of 0.0004. All employed lr_step, decaying by 0.5 every 10 steps. The integrating layer was a transformer; for Cifar100 and Tiny-ImageNet datasets, it consisted of 2 layers with 4 transformer heads each, and for Cifar10, 1 layer with 16 transformer heads. We observed that including optional steps in each node's training process resulted in better outcomes than excluding them. The training that included optional steps was conducted for 60 rounds, whereas the training without optional steps spanned 120 rounds. The results are as shown in Fig. 8. Therefore, subsequent experiments included these steps. We trained for 60 rounds, each involving 2 head trainings, 2 backbone trainings, and 2 full-layer trainings. The shared layer of the last client served as the desired shared feature extractor. We froze this shared layer and then trained a reinitialized head on all data (a mixture of all client data), obtaining the "Accuracy Indicative of Shared Layer Capability" on the test set after 20 epochs.

We then adopted the parameters of the last client's shared layer for all clients, froze these parameters, and fine-tuned each client's head 6 times to obtain personalized layers for different clients. Subsequently, we followed the method depicted in Fig. 5 (a) to acquire the "Global Model Accuracy."

The experimental results, as illustrated in Fig. 9, show that even under the most extreme conditions, i.e., in a pathological environment, the LI algorithm could achieve satisfactory results. Performance was slightly better at dir=0.01, improving further at dir=0.1, and when dir>=1 (dir>=0.1 for Cifar10), the LI algorithm achieved results very close to those obtained by training on a randomly mixed dataset of all client data. The dir=1 scenario already signifies severe data skew, as the dir value ranges from 0 to infinity. The 'Global Model Accuracy' and 'Accuracy Indicative of Shared Layer Capability' are very close. This fully demonstrates that simply stacking and then training a new network is an effective method for obtaining a global model. The 'Global Model Accuracy' and 'Accuracy Indicative of Shared Layer Capability' are very close when compared to "Baseline Accuracy with Combined Data". This demonstrates the effectiveness of the LI algorithm in extracting common features from heterogeneous datasets.

## 5 Conclusion and Prospect

In this study, we introduced the LI method, which combines end-to-end training with layer-wise training for federated learning and multi-task learning. We analyzed why it is effective and empirically validated the LI algorithm, demonstrating its promising results in the realms of federated learning and multi-task learning. A standout feature of the LI algorithm is its ability to effectively extract shared features across various tasks or clients, regardless of data heterogeneity. This surpasses current technical benchmarks in significant ways. Multi-task learning and federated personalized learning have produced results superior to those of individual task training, thereby giving the impression that training is conducted on a significantly larger dataset. For situations requiring a global model, we discovered that layering the output of each personalized layers and training an additional network can efficiently produce a global model.

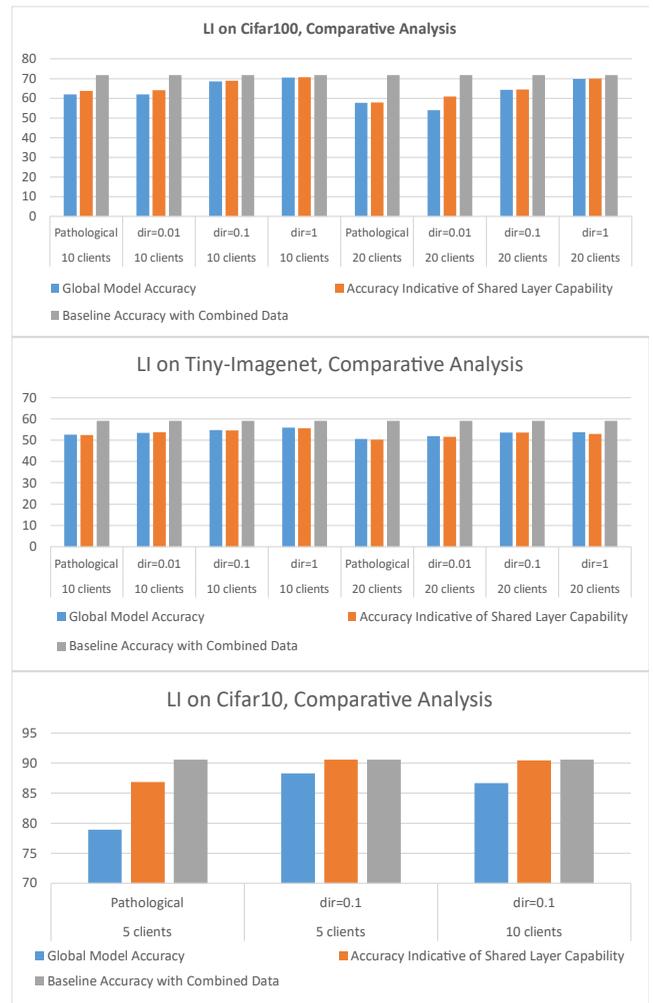

Fig. 9. Comparative Analysis of Different Settings: Global Model Accuracy, Shared Layer Capability Accuracy, and Baseline Accuracy with Combined Data Training. The values represent the percentage of accuracy.

However, several important questions remain to be addressed, such as how to properly partition shared and personalized layers, and whether there is a need to divide the neural network into more parts (three or more). These issues will be a primary focus in our future research. Although the training process appears to be sequential, in practice, as soon as the next client starts its training, the current client is also able to begin its training phase. This creates a structure like a looping pipeline. Nonetheless, this method does lead to increased communication costs, which must be carefully considered. Future research will focus on finding an optimal balance between parallel training efficiency and the associated communication overhead.

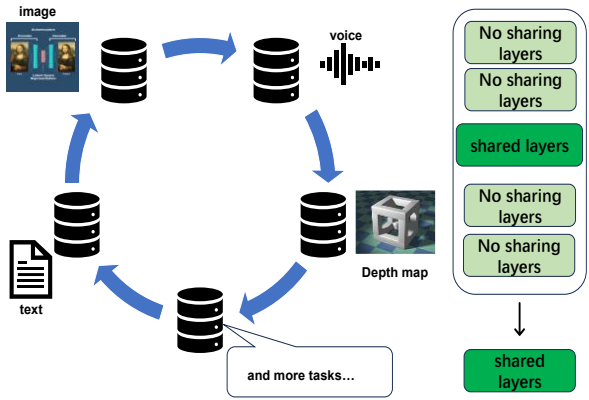

Fig. 10. LI Method applied to multi-modal data, extracting cross-modal shared features with the shared layer centrally positioned. This central placement is essential for effectively unifying diverse modalities in multi-modal datasets.

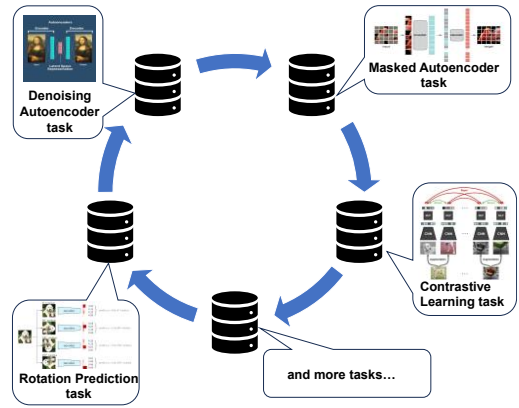

Fig. 11. Wider Application of the LI Method: Utilizing Various Task Types to Obtain More General Shared Features.

While the experiments in this study mainly considered situations with different label distributions, our theoretical analysis suggests that the LI algorithm may also be applicable to other non-IID scenarios. The LI method can be used to extract features from different datasets, even across modalities, as shown in Fig. 10. In such cases, the proportion of parameters in the shared layers should be particularly small, and perhaps the shared layers should be placed in the middle. Such shared layers, extracting abstract features, combined with the layer-wise training from high to low layers, might provide insights into understanding how the brain works. The LI method can also be applied across different types of tasks, as shown in Fig. 11, where the shared layers should be either at the bottom or very close to it. Here again, the proportion of shared layers is very small, making these layers as feature extractors with great generalizability. However, these assumptions still require further empirical validation. We look forward to extending the scope of the LI algorithm and exploring its efficacy in a broader range of application scenarios in future studies. Our work is preliminary, and we hope it serves to inspire further research.

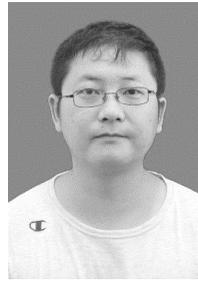

**Fei Li** is currently pursuing a Ph.D. in Artificial Intelligence at the University of Malaya. After completing his undergraduate education in 2005 at Sichuan Normal University, Chengdu, China. He graduated with a master's degree in computer science and technology from Zhejiang University, Hangzhou, China, in 2013. In 2015, he commenced his career at Neijiang Normal University, where he currently holds a position as a lecturer in the College of Artificial Intelligence. His research interests are focused on brain-inspired neural networks, continual learning, and federated learning.

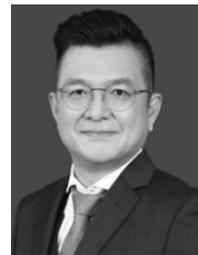

**Chu Kiong Loo** (Senior Member, IEEE) received the BEng First Class Hons in mechanical engineering from the University of Malaya and the PhD degree from University Sains Malaysia. He was a design engineer in various industrial firms and is the founder of the Advanced Robotics Lab. with the University of Malaya. He has been involved in the application of research into Perus's Quantum Associative Model and Pribram's Holonomic Brain Model in humanoid vision projects. Currently he is professor of Computer Science and Information Technology with the University of Malaya, Malaysia. He has led many projects funded by the Ministry of Science in Malaysia and the High Impact Research Grant from the Ministry of Higher Education, Malaysia. Loo's research experience includes brain-inspired quantum neural networks, constructivism-inspired neural networks, synergetic neural networks and humanoid research.

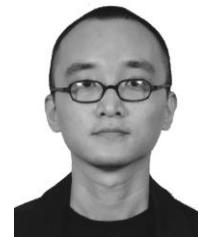

**Wei Shiung Liew** was born in Kuala Lumpur, Malaysia, in 1986. He received the B.S. degree in electronics engineering, majoring in robotics and automation from Multimedia University, Malaysia, in 2010, and the M.S. degree in biomedical engineering from the University of Malaya, in 2015. From 2009 to 2011, he was a Research Assistant with the Faculty of Information Science and Technology, Multimedia University. Since then, he has been a Postgraduate Research Assistant with the Department of Artificial Intelligence, University of Malaya. His research interests include evolutionary algorithms and biologically-inspired artificial intelligence methods.

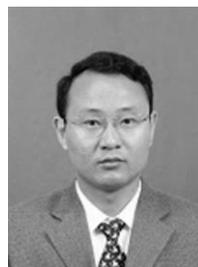

**Xiaofeng Liu** received the B.S. degree in electronic engineering and the M.S. degree in computer application from Taiyuan University of Technology, Taiyuan, China, in 1996 and 1999, respectively, and the Ph.D. degree in biomedical engineering from Xi'an Jiaotong University, Xi'an, China, in 2006. In 2006, he joined as an Associate Professor the College of Information and Electrical Engineering, Shandong University of Science and Technology. From 2008 to 2011, he worked as a Postdoc in the Institute of Artificial Intelligence and Robotics at Xi'an Jiaotong University. Since 2011, he has been with the Department of Telecommunication Engineering, Hohai University, Changzhou, where he is now a Professor and Director of the Laboratory of Cognition and Robotics. His current research interests focus on the study of neural information processing and nature-inspired navigation.